\documentclass[letterpaper,10pt,conference]{ieeeconf}

\IEEEoverridecommandlockouts
\usepackage[T1]{fontenc}
\usepackage{graphicx}
\usepackage{amsmath}
\usepackage{amssymb}
\usepackage{amsfonts}
\usepackage{booktabs}
\usepackage{pifont}
\newcommand{\cmark}{\ding{51}}
\newcommand{\xmark}{\ding{55}}
\usepackage{multirow}
\usepackage{microtype}
\usepackage{needspace}
\usepackage{xcolor}
\usepackage{cite}
\usepackage{url}
\usepackage[hidelinks]{hyperref}
\usepackage{cleveref}

\let\citep\cite

\newcommand{\pmark}{\(\triangle\)}

\title{\LARGE \bf
ViTacLab: Physically Consistent Visuo-Tactile Simulation, Benchmarking, and Representation Learning for Robotic Manipulation
}

\author{\parbox{\textwidth}{\centering
\fontsize{9.5}{13}\selectfont
Linjing You$^{1,2,3,*}$\hspace{0.9em}Zeming Yang$^{4,*}$\hspace{0.9em}Zhenhao Shen$^{4,*}$\hspace{0.9em}Shengqiang Xu$^{4,*}$\hspace{0.9em}Guanda Li$^{5}$\hspace{0.9em}Wenxuan Li$^{5}$ \\[2pt]Jiyu Guo$^{1}$\hspace{0.9em}Jasper Lu$^{4}$\hspace{0.9em}Kyle Xu$^{6}$\hspace{0.9em}Shawn Xie$^{6}$\hspace{0.9em}Junjie Zhu$^{5}$\hspace{0.9em}Siyu Wu$^{5}$ \\[2pt]Jian Liu$^{2}$\hspace{0.9em}Changhao Zhang$^{2}$\hspace{0.9em}Rongchong Li$^{7}$\hspace{0.9em}Zeyi Li$^{8}$\hspace{0.9em}Hengshuang Zhao$^{1}$\hspace{0.9em}Ruihai Wu$^{9,\dagger}$\\[7pt]
\fontsize{8}{10.5}\selectfont
$^1$The University of Hong Kong\enspace $^2$Ant Group\enspace $^3$Shenzhen Loop Area Institute (SLAI)\\[1pt]
$^4$Peking University\enspace $^5$Zhejiang University\\[1pt]
$^6$Lightwheel\enspace $^7$PsiBot\enspace $^8$National University of Singapore\enspace $^9$University of California, Berkeley\\[5pt]
$^*$Equal contribution.\qquad $^\dagger$Corresponding author: \texttt{ruihai@berkeley.edu}
}}

\begin{document}
\raggedbottom
\maketitle
\thispagestyle{empty}
\pagestyle{empty}

\begin{abstract}
Tactile sensing is essential for contact-rich robotic manipulation, yet existing
visuo-tactile frameworks face a persistent trade-off between physically faithful
but non-scalable simulators and GPU-parallel approximations with inconsistent
contact responses. Existing tactile manipulation benchmarks further remain
largely restricted to two-finger gripper tasks, limiting evaluation across
diverse embodiments and contact regimes. We present \emph{ViTacLab}, a
full-stack visuo-tactile framework comprising \emph{ViTacSim}, \emph{ViTacBench},
and \emph{ViTacPolicy}. \emph{ViTacSim} introduces a tri-modal tactile primitive consisting of a global sparse contact-force lattice, a local dense force array, and tactile images. Anchored to contact-solver outputs, it corrects normal inconsistency via local stiffness alignment and reconstructs tangential force fields under decoupled slip--stick regimes while preserving GPU parallelism.
\emph{ViTacBench} establishes a hierarchical
benchmark spanning parallel grippers, dexterous hands, and bimanual systems,
with tasks graduated from vision-dominant to tactile-dominant regimes to
systematically diagnose the modality shift from vision to touch.
\emph{ViTacPolicy} aggregates an arbitrary number of heterogeneous tactile
inputs via spatial-pose embeddings and a Transformer module, and aligns dense
force-array and tactile-image modalities into a unified discrete representation
through a two-stream VQ-VAE paradigm with a shared quantized codebook. A joint
projection preserves the sparse PhysX wrench while enforcing local friction
cones.
Controlled Hertzian-reference indentation and real-sensor sliding tests reduce
normal and tangential force RMSE from 2.31 and 1.85~N to 0.42 and 0.53~N,
respectively, while runtime reaches 2,482.9 tactile-pipeline FPS at 256 parallel
environments. Simulation and
real-world results show gains on several contact-rich tasks, while low absolute
success on tactile-dominant tasks exposes the remaining difficulty of purely
tactile control.
\end{abstract}

\section{Introduction}
\label{sec:intro}

\begin{figure*}[t]
\centering
\includegraphics[width=\textwidth]{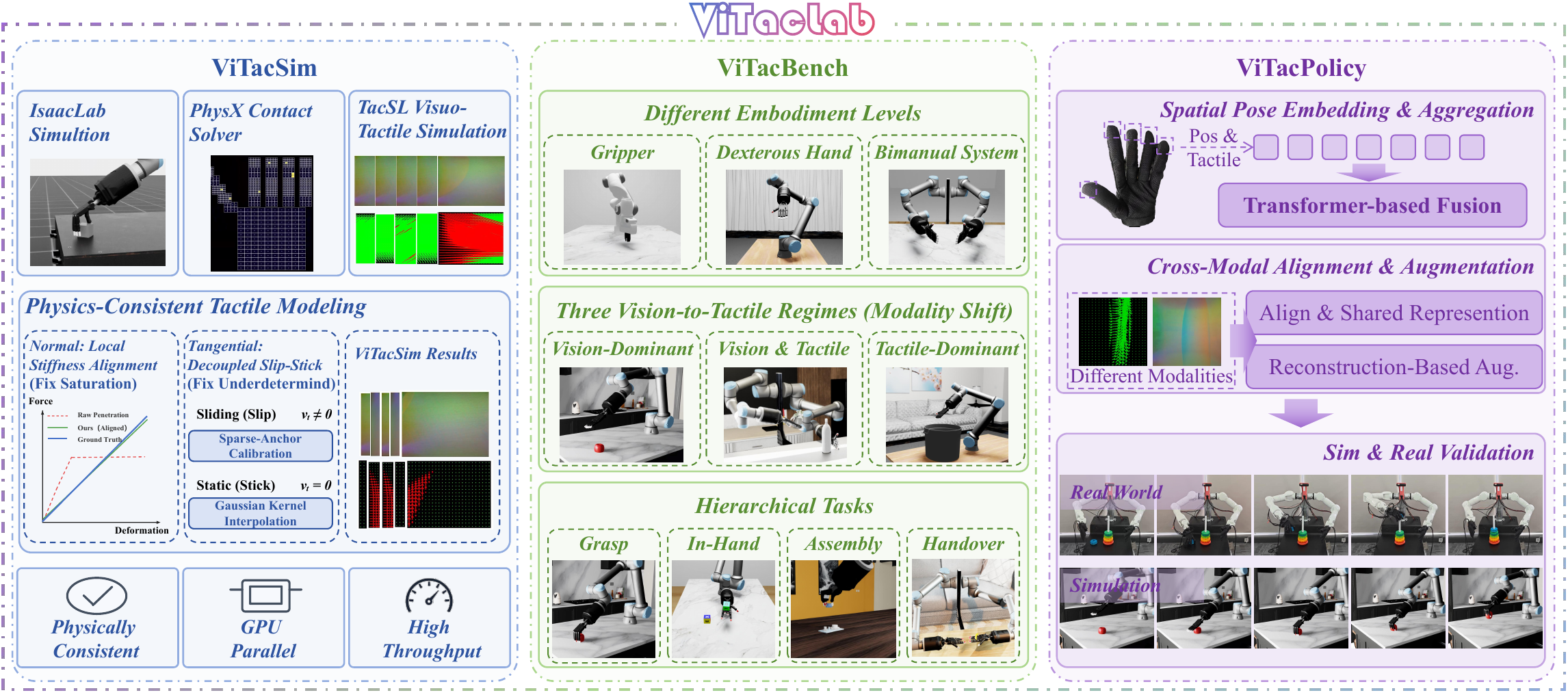}
\caption{
\textbf{Overview of ViTacLab.} ViTacLab unifies three components:
\textbf{ViTacSim}, a contact-solver-anchored, GPU-parallel tactile simulator;
\textbf{ViTacBench}, a hierarchical benchmark spanning parallel grippers,
dexterous hands, and bimanual systems across vision-to-touch regimes; and
\textbf{ViTacPolicy}, a sensor-agnostic policy that fuses heterogeneous tactile
inputs and aligns dense force array and tactile image modalities via
shared-codebook VQ-VAE pretraining.
}
\vspace{-0.5em}
\label{fig:overview}
\end{figure*}

Robotic manipulation in unstructured environments is fundamentally
contact-rich. Recent large-scale robot learning has driven rapid progress in
vision-centric manipulation, enabled by scalable visual policies,
standardized benchmarks, and large-scale real and synthetic
datasets~\citep{mandlekar2021matterslearningofflinehuman,zhao2023learningfinegrainedbimanualmanipulation,chi2024diffusionpolicyvisuomotorpolicy}.
However, when the robot occludes objects or tasks require sub-centimeter
precision, vision degrades and latent contact mechanics become the primary
source of state information. Tactile sensing offers a direct window into these
interactions~\citep{8403291,lee2019makingsensevisiontouch}, but tactile-driven robot
learning has not shown the same full-stack compounding progress.

This stagnation stems from three unresolved technical challenges.
\textbf{(1) Simulation: fidelity vs.\ efficiency.}
GPU-parallel simulation methods~\citep{10912733, nguyen2024tacexgelsighttactilesimulation} typically derive tactile outputs from raw
penetration depth, causing two physical failure modes: normal responses saturate
at solver-imposed depth limits, breaking deformation--force consistency, while
tangential forces become underdetermined as relative velocity vanishes in
static-friction regions. High-fidelity simulators address these issues through
explicit contact mechanics~\citep{Li2020IPC,li2025taccelscalingvisionbasedtactile}, but are too
costly for large-scale policy learning. \textbf{(2) Evaluation: narrow embodiments and undiagnosed modality shift.}
Most existing tactile manipulation benchmarks focus on parallel grippers and
isolated vision- or touch-dominant regimes~\citep{chen2026univtacunifiedsimulationplatform, li2024maniskillvitac2025challengemanipulation, luu2026manifeelbenchmarkingunderstandingvisuotactile}, leaving dexterous hands and bimanual
systems underexplored. As a result, they cannot systematically evaluate how
tactile feedback scales with embodiment complexity, nor answer a central
question in dexterous manipulation: as visual context degrades, when and how
does touch become the dominant control modality? \textbf{(3) Architecture: rigid configurations and unaligned modalities.}
Standard visuo-tactile policies are often tied to fixed sensor counts and
placements, and fuse heterogeneous dense force-array and tactile-image signals
by simple concatenation, without enforcing physical alignment across
modalities~\citep{heng2025vitacformerlearningcrossmodalrepresentation,
huang2025tactilevlaunlockingvisionlanguageactionmodels,
huang20253dvitaclearningfinegrainedmanipulation}.

To address these bottlenecks and establish a full-stack foundation for tactile robot learning, we present \textbf{ViTacLab}, a unified framework comprising three components.

\Needspace{3\baselineskip}\noindent\textbf{1. ViTacSim} generates tri-modal tactile outputs by anchoring them to PhysX contact-solver outputs~\citep{makoviychuk2021isaac} rather than raw penetration depth. Its tactile primitive comprises a sparse contact-force lattice, dense force arrays, and tactile images. Through local stiffness alignment, decoupled slip--stick reconstruction, and a global wrench projection, ViTacSim corrects normal and tangential inconsistencies while preserving the sparse solver wrench and GPU parallelism. We validate both force components directly and report complete 1--256-environment throughput and memory scaling.

\Needspace{3\baselineskip}\noindent\textbf{2. ViTacBench} establishes a hierarchical tactile manipulation benchmark spanning parallel grippers, dexterous hands, and bimanual systems, with tasks covering vision-dominant, visuo-tactile, and tactile-dominant regimes.

\Needspace{3\baselineskip}\noindent\textbf{3. ViTacPolicy} supports heterogeneous tactile inputs through spatial-pose embeddings and Transformer-based fusion~\citep{vaswani2023attentionneed}, and aligns dense force-array and tactile-image modalities into a unified discrete representation via a shared-codebook VQ-VAE~\citep{oord2018neuraldiscreterepresentationlearning} pre-training paradigm.

Together, these three pillars are evaluated in simulation and on real hardware.
The results establish direct normal/tangential force improvements, solver-wrench
consistency by construction, complete batched runtime and memory scaling, and
task-dependent manipulation gains. A 50-trial-per-task transfer study and
matched unimodal controls separate aggregate improvements from the remaining
per-task uncertainty, while higher-data Level~3 experiments show continued
scaling with demonstrations.

\section{Related Work}

\paragraph{Tactile Simulation.}

Existing tactile simulators still trade contact fidelity for policy-learning throughput. Rendering-based systems such as TACTO~\citep{Wang_2022} and Taxim~\citep{si2021taximexamplebasedsimulationmodel} efficiently synthesize tactile images, but rely on simplified mappings rather than physically grounded force fields. GPU-parallel systems such as TacSL~\citep{10912733} and TacEx~\citep{nguyen2024tacexgelsighttactilesimulation} integrate tactile rendering or force fields into Isaac Gym~\citep{makoviychuk2021isaac} / Isaac Lab~\citep{mittal2025isaaclab}, yet still rely on penetration or velocity cues, leading to inconsistent normal and tangential responses. High-fidelity soft-body simulators such as Taccel~\citep{li2025taccelscalingvisionbasedtactile} improve realism through explicit contact mechanics, but remain computationally expensive. In contrast, \emph{ViTacSim} anchors tactile generation to PhysX solver contacts, producing $T_{\text{sparse}}$, $T_{\text{dense}}$, and $T_{\text{img}}$ while correcting normal inconsistency and reconstructing tangential fields under decoupled slip--stick regimes.

\paragraph{Tactile Benchmarks.} Recent tactile benchmarks have advanced perception and manipulation evaluation,
but remain limited in embodiment coverage and modality diagnosis. Perception
benchmarks such as Tactile MNIST~\citep{schneider2025tactilemnistbenchmarkingactive} study active
tactile recognition tasks, while manipulation benchmarks such as
ManiSkill-ViTac~\citep{li2024maniskillvitac2025challengemanipulation}, VTDexManip~\citep{liu2025vtdexmanip}, ManiFeEL~\citep{luu2026manifeelbenchmarkingunderstandingvisuotactile}, and UniVTac~\citep{chen2026univtacunifiedsimulationplatform} evaluate tactile or visuo-tactile control under specific
platforms, task suites, or sensing settings. However, they do not systematically
span parallel grippers, dexterous hands, and bimanual systems, nor organize
tasks by graduated vision-to-touch dependence. \emph{ViTacBench} fills this gap
by jointly varying embodiment complexity and modality regime.

\paragraph{Tactile Encoders and Policies.} Recent tactile encoders target
cross-sensor and vision/language alignment~\citep{yu2024octopiobjectpropertyreasoning,xu2025unit,yang2024unitouch,feng2025anytouchlearningunifiedstaticdynamic},
while manipulation policies fuse visual and tactile observations for
contact-rich control~\citep{xue2025reactive,heng2025vitacformerlearningcrossmodalrepresentation,huang2025tactilevlaunlockingvisionlanguageactionmodels,huang20253dvitaclearningfinegrainedmanipulation}.
Most remain tied to fixed sensor formats, platforms, or layouts. ViTacPolicy
instead accepts heterogeneous $T_{\text{dense}}$ and $T_{\text{img}}$ inputs,
pretrains their encoders jointly, and adapts to different tactile layouts.

\begin{table*}[t]
\centering
\caption{Comparison with existing tactile simulation, benchmark, and encoder works.}
\label{tab:comparison}
\resizebox{\textwidth}{!}{
\begin{tabular}{llcc|ccccc|cc}
\toprule
& & \multicolumn{2}{c|}{\textbf{Simulation}}
& \multicolumn{5}{c|}{\textbf{Benchmark}}
& \multicolumn{2}{c}{\textbf{Encoder}} \\
\textbf{Category} & \textbf{Method}
& \textbf{Phys.\ Consist.} & \textbf{GPU-Par.}
& \textbf{$T_{\text{sparse}}$} & \textbf{$T_{\text{dense}}$}
& \textbf{$T_{\text{img}}$}
& \textbf{DexHand/Bimanual} & \textbf{Mod.\ Shift}
& \textbf{Flex.\ Layout} & \textbf{Dense-Img Align.} \\
\midrule
\multirow{4}{*}{Simulation}
& TACTO~\citep{Wang_2022}    & \xmark & \xmark & {--} & {--} & {--} & {--} & {--} & {--} & {--} \\
& TacSL~\citep{10912733} & \xmark & \cmark & {--} & {--} & {--} & {--} & {--} & {--} & {--} \\
& Taccel~\citep{li2025taccelscalingvisionbasedtactile}    & \cmark & \cmark & {--} & {--} & {--} & {--} & {--} & {--} & {--} \\
& TacEx~\citep{nguyen2024tacexgelsighttactilesimulation}  & \pmark & \pmark & {--} & {--} & {--} & {--} & {--} & {--} & {--} \\
\midrule
\multirow{4}{*}{Benchmark}
& ManiSkill-ViTac~\citep{li2024maniskillvitac2025challengemanipulation} & {--} & {--} & \xmark & \xmark & \cmark & \xmark & \pmark & {--} & {--} \\
& VTDexManip~\citep{liu2025vtdexmanip}         & {--} & {--} & \pmark & \xmark & \xmark & \cmark & \xmark & {--} & {--} \\
& ManiFeel~\citep{luu2026manifeelbenchmarkingunderstandingvisuotactile}             & {--} & {--} & \xmark & \cmark & \cmark & \xmark & \pmark & {--} & {--} \\
& UniVTac~\citep{chen2026univtacunifiedsimulationplatform}              & {--} & {--} & \xmark & \cmark & \cmark & \xmark & \pmark & \xmark & \xmark \\
\midrule
\multirow{2}{*}{Encoder}
& AnyTouch~\citep{feng2025anytouchlearningunifiedstaticdynamic} & {--} & {--} & {--} & {--} & {--} & {--} & {--} & \xmark & \xmark \\
& UniTouch~\citep{yang2024unitouch} & {--} & {--} & {--} & {--} & {--} & {--} & {--} & \xmark & \xmark \\
\midrule
\textbf{Ours}
& \textbf{ViTacLab}
& \cmark & \cmark & \cmark & \cmark & \cmark & \cmark & \cmark & \cmark & \cmark \\
\bottomrule
\end{tabular}}
\par\vspace{4pt}
\begin{minipage}{\textwidth}
\fontsize{7.5}{9}\selectfont
\textbf{Phys.\ Consist.} denotes physically grounded normal/tangential contact modeling; \textbf{GPU-Par.} denotes GPU-parallel multi-environment simulation. Benchmark modality columns indicate support for sparse contact lattices $T_{\text{sparse}}$, dense force arrays $T_{\text{dense}}$, and tactile images $T_{\text{img}}$. \textbf{DexHand/Bimanual} denotes dexterous-hand or bimanual embodiments; \textbf{Mod.\ Shift} denotes graduated vision-to-touch regimes. \textbf{Flex.\ Layout} denotes flexible sensor count and placement; \textbf{Dense-Img Align.} denotes alignment between $T_{\text{dense}}$ and $T_{\text{img}}$. \pmark denotes partial support; {--} denotes not applicable.
\end{minipage}
\end{table*}

\section{ViTacSim: Physically Consistent Visuo-Tactile Simulation}
\label{sec:vitacsim}

\begin{figure*}[t]
\centering
\includegraphics[width=0.90\textwidth]{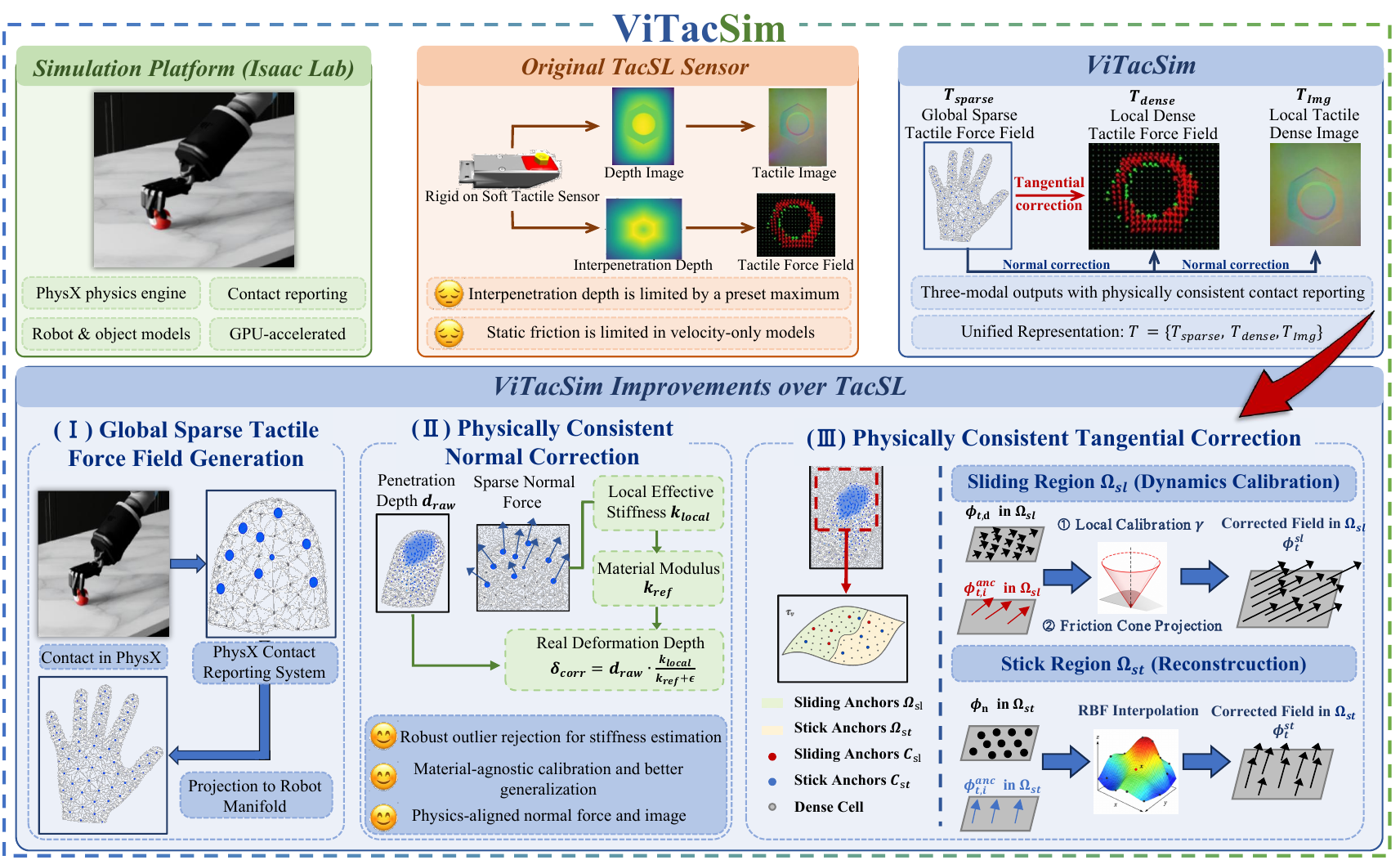}
\vspace{-0.5em}
\caption{ViTacSim unifies a global sparse contact-force lattice, local dense force array, and local tactile images through normal correction
and tangential reconstruction.}
\vspace{-0.5em}
\label{fig:sim}
\end{figure*}

As shown in \cref{fig:sim}, \emph{ViTacSim} constructs a tri-modal tactile
representation $\mathcal{T}$ consisting of a global sparse contact-force lattice
$T_{\text{sparse}}$, a local dense force array $T_{\text{dense}}$, and local
tactile images $T_{\text{img}}$, capturing both global contact topology and
local high-frequency tactile responses.

\paragraph{Tri-modal tactile representation.}
At each simulation step, we extract sparse solver contacts
$\mathcal{C}=\{(p_k,f_k,n_k,d_k,v_{t,k})\}_{k=1}^{N_c}$ from
PhysX~\citep{makoviychuk2021isaac}, where $p_k$, $f_k$, $n_k$, $d_k$, and
$v_{t,k}$ denote contact position, force, normal, raw penetration depth, and
relative tangential velocity. Projecting these contacts onto the tactile
manifold gives $T_{\text{sparse}}=\mathcal{P}(\mathcal{C})$, which captures
global contact topology but lacks local resolution. For each active patch
$\Omega$, ViTacSim constructs
$T_{\text{dense}}=\{\phi_n(x),\phi_t(x)\}_{x\in\Omega}$, where $\phi_n(x)$ is
the scalar normal force field and $\phi_t(x)\in\mathbb{R}^2$ is the tangential
force field in the local tangent basis.

\Needspace{6\baselineskip}
\paragraph{Normal correction.}
Raw solver penetration $d_{\text{raw}}(x)$ is not elastomer deformation, so the
normal response saturates at high loads:
\begin{equation}
\label{eq:normal-correction}
\begin{aligned}
k_{\text{local}}(x)
&=\operatorname*{RobustMean}_{i\in\mathcal{N}(x)}
\left(\frac{F_{n,i}}{d_i+\epsilon}\right), \\
F_{n,i}&=\max(f_i^\top n_i,0), \\
\delta_{\text{corr}}(x)
&=d_{\text{raw}}(x)
\frac{k_{\text{local}}(x)}{k_{\text{ref}}+\epsilon},\\
\phi_n(x)&=k_{\text{ref}}\delta_{\text{corr}}(x).
\end{aligned}
\end{equation}

Here $\mathcal{N}(x)$ is the neighboring sparse-contact set, and $n_i$ points
along positive compression.

This correction drives both the normal force array and the deformation map; tactile
images use standard optical rendering models~\citep{Wang_2022,si2021taximexamplebasedsimulationmodel}.
We update $k_{\text{ref}}$ as the running median of
valid solver-force-to-depth ratios $F_{n,i}/(d_i+\epsilon)$.

\paragraph{Tangential reconstruction.}
We decompose each patch into sliding and sticking regions by tangential
velocity:
\begin{equation}
\label{eq:slip-stick-partition}
\Omega_{\mathrm{sl}}
=
\{x\in\Omega:\|v_t(x)\|>\tau_v\},
\qquad
\Omega_{\mathrm{st}}
=
\Omega\setminus\Omega_{\mathrm{sl}} .
\end{equation}
Here $v_t(x)$ is the local tangential velocity assigned to dense cell $x$ from
the corresponding projected solver contact, and $\tau_v$ is obtained from the
Coulomb-cone slip criterion. Let $\mathcal{C}_{\mathrm{sl}}$ and
$\mathcal{C}_{\mathrm{st}}$ denote sparse anchors in the sliding and sticking
regions, with projected locations $c_i$ and tangential anchor forces
$\phi_{t,i}^{\mathrm{anc}}$.

For sliding contacts, dense tangential predictions $\phi_{t,d}$ are calibrated
by sparse anchors and projected onto the Coulomb friction disk:
\begin{equation}
\label{eq:sliding-tangential}
\begin{aligned}
\phi_t^{\mathrm{sl}}(x)
&=\Pi_{\mu\phi_n(x)}
\left(\gamma(x)\phi_{t,d}(x)\right), \\
\gamma(x)
&=\operatorname*{RobustMean}_{i\in\mathcal{N}(x)\cap\mathcal{C}_{\mathrm{sl}}} r_i,
\\
r_i&=\frac{\|\phi_{t,i}^{\mathrm{anc}}\|}
{\|\phi_{t,d}(c_i)\|+\epsilon}.
\end{aligned}
\end{equation}
where $\Pi_{\rho}(\cdot)$ projects a 2D tangential vector onto the friction disk
with radius $\rho$.

For sticking contacts, where velocity-based prediction is underdetermined, we
interpolate sparse static anchors and enforce the same friction bound:
\begin{equation}
\label{eq:sticking-tangential}
\begin{aligned}
\phi_t^{\mathrm{st}}(x)
&=\Pi_{\mu\phi_n(x)}
\left(\sum_{i\in\mathcal{C}_{\mathrm{st}}}
w_i(x)\phi_{t,i}^{\mathrm{anc}}\right), \\
w_i(x)
&=
\frac{\exp(-\|x-c_i\|^2/\sigma^2)}
{\sum_{j\in\mathcal{C}_{\mathrm{st}}}\exp(-\|x-c_j\|^2/\sigma^2)}
.
\end{aligned}
\end{equation}
We set the Gaussian bandwidth $\sigma$ to $1.5$ times the contact radius.
All operations are local projection, interpolation, and scatter steps,
preserving GPU-parallel execution across environments. The final tangential field is
\begin{equation}
\label{eq:unified-tangential}
\phi_t(x)=
\begin{cases}
\phi_t^{\mathrm{sl}}(x), & x\in\Omega_{\mathrm{sl}},\\
\phi_t^{\mathrm{st}}(x), & x\in\Omega_{\mathrm{st}} .
\end{cases}
\end{equation}
Let $\widetilde{\phi}$ collect the reconstructed normal and tangential taxel
forces, and let $A$ map them to the sensor-frame force and moment about $p_0$.
The final projection jointly enforces the sparse-contact wrench and every local
friction cone:
\begin{equation}
\label{eq:wrench-projection}
\begin{aligned}
\phi^\star&=\arg\min_{\phi\in\mathcal{K}}
\|\phi-\widetilde{\phi}\|_2^2
\quad\mathrm{s.t.}\quad A\phi=w_{\mathrm{PhysX}},\\
w_{\mathrm{PhysX}}&=\sum_i
\begin{bmatrix}f_i\\(p_i-p_0)\times f_i\end{bmatrix},\\
\mathcal{K}&=\{\phi:\phi_{n,j}\geq0,\
\|\phi_{t,j}\|_2\leq\mu\phi_{n,j},\ \forall j\}.
\end{aligned}
\end{equation}
Thus the integrated dense field preserves the sparse PhysX force and torque
without allowing the final projection to violate local Coulomb bounds.

\section{ViTacBench: Modality-Shift Benchmark}
\label{sec:benchmark}

\begin{figure*}[t]
\centering
\includegraphics[width=0.9\textwidth]{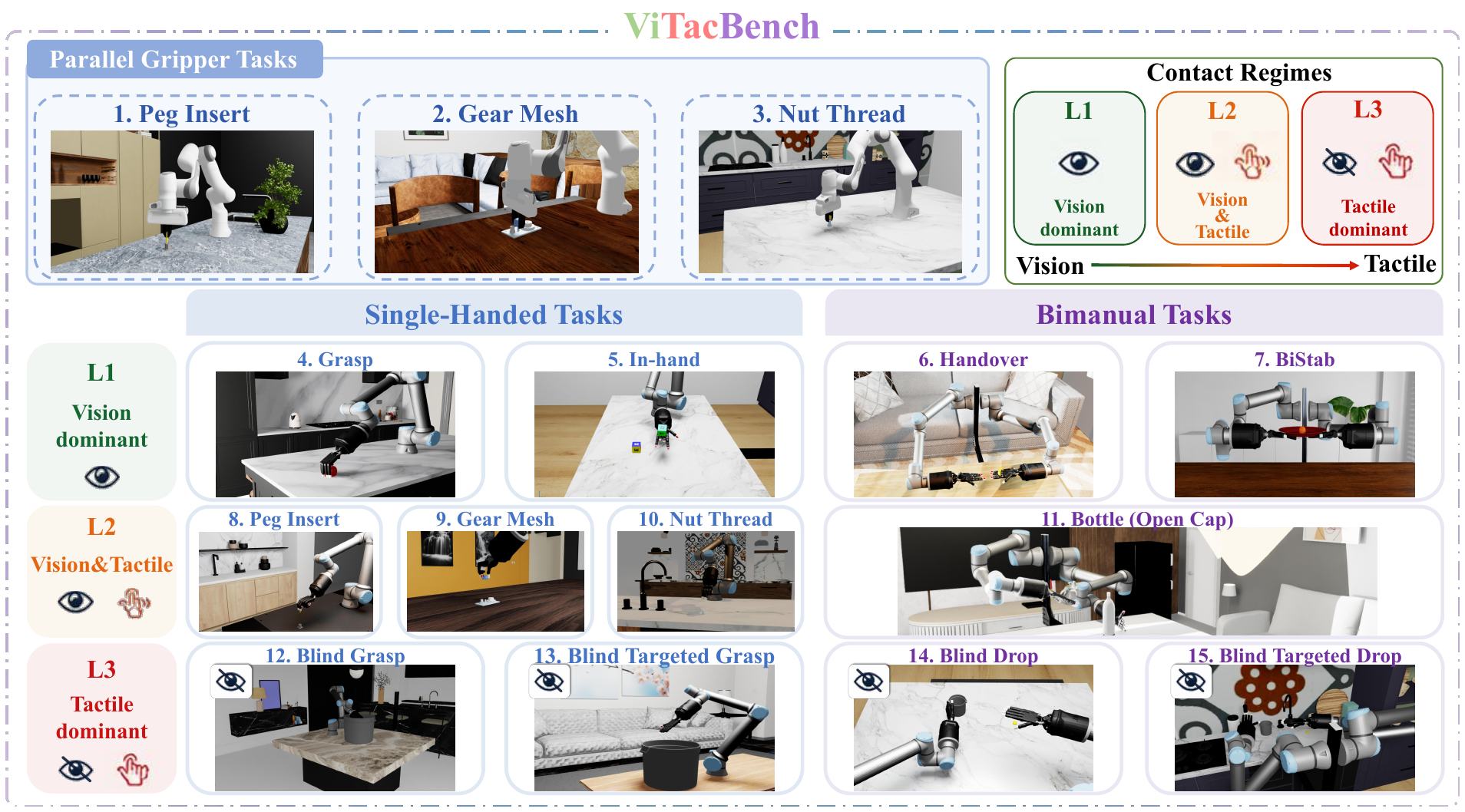}
\caption{ViTacBench spans parallel-gripper, dexterous-hand, and bimanual tasks
across vision-dominant, visuo-tactile cooperative, and tactile-dominant
regimes.}
\vspace{-0.5em}
\label{fig:bench}
\end{figure*}

As shown in \cref{fig:bench}, ViTacBench evaluates tactile perception and
multimodal fusion along two axes: robot embodiment and modality dependence. It
spans parallel grippers, dexterous hands, and bimanual systems, with tasks
progressively shifting reliance from vision to touch.

\textbf{Parallel gripper tasks} include peg insertion, gear meshing, and nut
threading, providing controlled low-DoF testbeds for contact-rich alignment.
\textbf{Dexterous hand tasks} include tabletop grasping, in-hand manipulation,
bi-stable transfer, dexterous peg insertion, blind grasping, and targeted
grasping, evaluating tactile feedback for contact localization and fine-grained
control. \textbf{Bimanual tasks} include handover, bottle opening, object
twisting, low-light coordinated manipulation, and bimanual writing, requiring
cross-arm coordination and sustained contact regulation.

Tasks are further grouped into \textbf{vision-dominant},
\textbf{visuo-tactile cooperative}, and \textbf{tactile-dominant} regimes,
depending on whether visual observations provide sufficient state information,
tactile feedback is needed for precise contact control, or visual context is
degraded such that success primarily depends on touch. The 15-task suite is
summarized in \cref{fig:bench}. In Level~3, wrist-camera observations are fully
masked and the remaining global view is occluded by design; policies therefore
receive no direct wrist-view object pixels.

\section{ViTacPolicy: Visuo-Tactile Policy Learning}
\label{sec:vitacpolicy}

\begin{figure*}[t]
\centering
\includegraphics[width=0.9\textwidth]{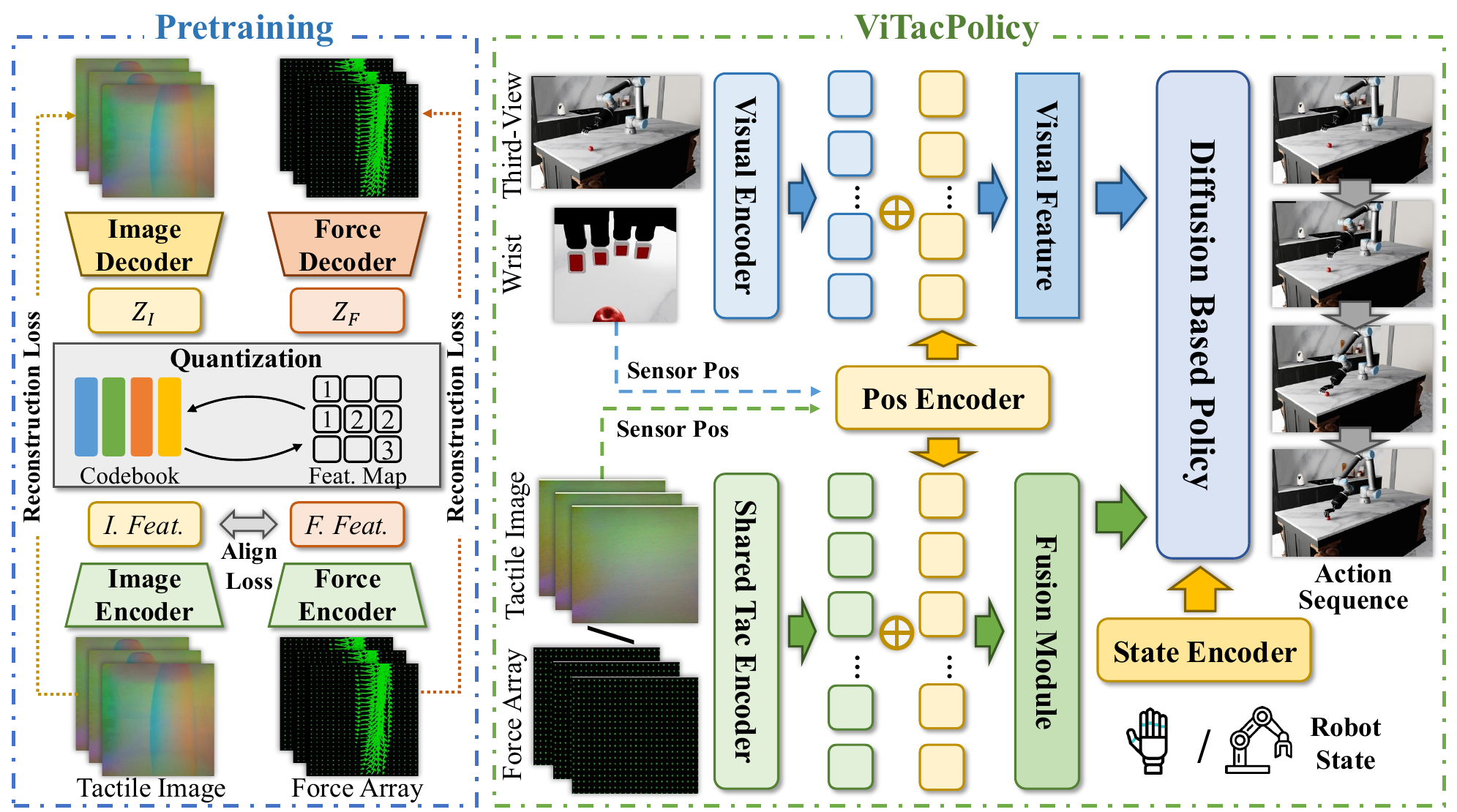}
\caption{ViTacPolicy aggregates a variable number of spatially distributed
tactile inputs through pose-conditioned Transformer fusion, and aligns
$T_{\text{dense}}$ and $T_{\text{img}}$ modalities via shared-codebook VQ-VAE
pre-training.}
\vspace{-0.5em}
\label{fig:policy}
\end{figure*}

\subsection{Spatial-Pose-Conditioned Multimodal Aggregation}

ViTacPolicy handles heterogeneous tactile sensors with variable counts and
placements by encoding each tactile input together with its spatial pose. Let
$\{T_{i,m_i}\}_{i=1}^{N_{\text{tac}}}$ denote the tactile observations, where
$m_i\in\mathcal{M}_{\text{tac}}=\{\text{dense},\text{img}\}$ indicates whether
the input is a dense force array from $T_{\text{dense}}$ or an optical tactile
image from $T_{\text{img}}$. A modality-specific tactile encoder extracts
$h_i=E_{m_i}(T_{i,m_i})\in\mathbb{R}^{d_h}$, using ResNet-style encoders for
image-like tactile inputs~\citep{he2015deepresiduallearningimage}. Each sensor pose
$X_i\in\mathbb{R}^{9}$, represented by 3D position and 6D orientation, is mapped
to a pose embedding $P_i=\mathrm{MLP}(X_i)\in\mathbb{R}^{d_h}$ and added to the
tactile feature, giving the pose-aware token $Z_i=h_i+P_i$. The resulting tokens are aggregated by a Transformer encoder~\citep{vaswani2023attentionneed}, followed by a pooling operation that produces the final tactile representation $h_{\text{tac}}$:
\begin{equation}
\label{eq:policy-aggregation}
\begin{aligned}
h_{\text{tac}}
&=\mathrm{Pool}\!\left(
\mathrm{Transformer}(\{Z_i\}_{i=1}^{N_{\text{tac}}})
\right), \\
a_t&\sim\pi_\theta(a_t\mid h_{\text{rgb}},h_{\text{tac}},q_t).
\end{aligned}
\end{equation}
By conditioning the policy on the fused tactile representation $h_{\text{tac}}$,
RGB feature $h_{\text{rgb}}$, and proprioceptive state $q_t$, this aggregation
decouples policy learning from fixed tactile layouts and supports deployment
across grippers, dexterous hands, and bimanual systems.

\subsection{Shared-Codebook Two-Stream Tactile Pretraining}

To learn modality-consistent tactile representations, we pretrain the tactile
encoder with a shared-codebook two-stream VQ-VAE objective~\citep{
oord2018neuraldiscreterepresentationlearning}. We consider paired or
co-located tactile observations from two local tactile modalities,
$\mathcal{M}_{\text{tac}}=\{\text{dense},\text{img}\}$, where
$T_{i,\text{dense}}$ denotes a dense force array sample and
$T_{i,\text{img}}$ denotes an optical tactile image sample. We denote a
mini-batch of paired tactile samples as
$\mathcal{B}=\{(T_{i,\text{dense}},T_{i,\text{img}})\}_{i=1}^{|\mathcal{B}|}$.
Each modality uses a modality-specific encoder-decoder pair, while both
streams share a discrete codebook
$\mathcal{C}_{\mathrm{vq}}=\{c_k\}_{k=1}^{K}$.

Let $E_m^{\mathrm{map}}$ denote the encoder output before the final pooling
layer. For modality $m\in\mathcal{M}_{\text{tac}}$, we quantize the
pre-pooling tactile feature map, rather than the final global embedding:
\begin{equation}
\label{eq:vq-feature-map}
\begin{aligned}
F_{i,m}
&=E_m^{\mathrm{map}}(T_{i,m}), \\
\hat{F}_{i,m}
&=Q(F_{i,m};\mathcal{C}_{\mathrm{vq}}), \\
\hat{T}_{i,m}
&=D_m(\hat{F}_{i,m}).
\end{aligned}
\end{equation}
where $Q(\cdot;\mathcal{C}_{\mathrm{vq}})$ performs vector-wise quantization
over the spatial or token locations of the feature map using the shared
codebook.

The VQ-VAE loss combines modality-specific reconstruction with the standard
codebook and commitment losses on the quantized feature maps:
\begin{equation}
\label{eq:vq-loss}
\begin{aligned}
\mathcal{L}_{\mathrm{vq}}
&=\frac{1}{|\mathcal{B}|}
\sum_{i\in\mathcal{B}}
\sum_{m\in\mathcal{M}_{\text{tac}}}
\\[-0.2em]
&\quad{}
\left[
\left\|T_{i,m}-\hat{T}_{i,m}\right\|_2^2
+
\ell_{\mathrm{vq}}(F_{i,m},\hat{F}_{i,m})
\right],
\end{aligned}
\end{equation}
where $\ell_{\mathrm{vq}}$ denotes the standard VQ-VAE codebook and
commitment regularization terms.

In parallel, we align the global tactile representations produced after the
final pooling layer, where
$h_{i,m}=E_m(T_{i,m})=\mathrm{Pool}(F_{i,m})\in\mathbb{R}^{d_h}$ for
$m\in\mathcal{M}_{\text{tac}}$. For paired dense force-array and optical
tactile-image observations, we minimize the mean-squared distance between their
global representations and combine it with the VQ-VAE objective:
\begin{equation}
\label{eq:pretrain-loss}
\mathcal{L}_{\mathrm{pre}}
=
\mathcal{L}_{\mathrm{vq}}
+
\lambda_{\mathrm{align}}
\frac{1}{|\mathcal{B}|}
\sum_{i\in\mathcal{B}}
\left\|
h_{i,\text{dense}}
-
h_{i,\text{img}}
\right\|_2^2 .
\end{equation}

This design uses the shared codebook to learn local contact primitives at the
feature-map level, while the alignment loss encourages dense force arrays and
tactile images to share a common global tactile embedding space. The pretrained
encoders are then used to initialize ViTacPolicy for downstream visuo-tactile
manipulation.

\section{Experiments}
\label{sec:experiments}

We evaluate ViTacLab from four perspectives: \textbf{Q1:} Does ViTacSim improve
force accuracy against independent references, preserve the solver wrench, and
scale across GPU environments? \textbf{Q2:} Does
ViTacPolicy leverage touch for contact-rich manipulation? \textbf{Q3:} Does
ViTacBench capture the shift from vision to touch? \textbf{Q4:} Do
pose-conditioned encoding and shared-codebook pretraining improve modality
generalization and sim-to-real transfer?

\subsection{Setup}

\textbf{Real-World Setup.}
We evaluate ViTacLab in both parallel gripper and bimanual dexterous
settings, with three real-world tasks designed for each setup. For each task,
we collect 50 demonstration trajectories for training. After training, each
condition in the quantitative gripper study is evaluated with 50 trials per
task; the other physical deployments are evaluated qualitatively.

\textbf{Tasks and Data.}
We evaluate on the full ViTacBench suite spanning parallel grippers, dexterous
hands, and bimanual systems across vision-dominant, visuo-tactile cooperative,
and tactile-dominant regimes. For each task, we collect 100
trajectories for downstream policy learning and 100 additional trajectories for representation pretraining. Each policy
is trained with five random seeds and evaluated over 200 simulation episodes
per seed; tables report mean success rate as the primary metric.

\textbf{Simulation Setup.}
Physics runs at 120~Hz and control/rendering at 60~Hz. Across embodiments,
position/velocity solver iterations span $8/0$--$192/1$, contact/rest offsets
use $5/{-}2$ or $1/{-}1$~mm, and nominal friction spans $0.5$--$2.0$. These are
the settings used in the reported benchmark and do not establish parameter
robustness.

\textbf{Baselines.}
We compare against three state-of-the-art baselines that cover different input
modalities and policy designs: \textbf{DP}~\citep{chi2024diffusionpolicyvisuomotorpolicy}, \textbf{ADM\_DP}~\citep{wang2026admdpadaptivedynamicmodality} and \textbf{ViTacFormer}~\citep{heng2025vitacformerlearningcrossmodalrepresentation}. We also include
\textbf{Ours w/o PT}, a variant of ViTacPolicy without shared-codebook VQ-VAE
pretraining, to evaluate the contribution of the proposed tactile pretraining
module. A dash in a result table denotes an unevaluated combination whose
released baseline does not support the required embodiment or observation
interface; it is not counted as a failed trial.

\subsection{ViTacSim Force Fidelity, Wrench Consistency, and Scalability}

\begin{figure*}[t]
\centering
\includegraphics[width=\textwidth]{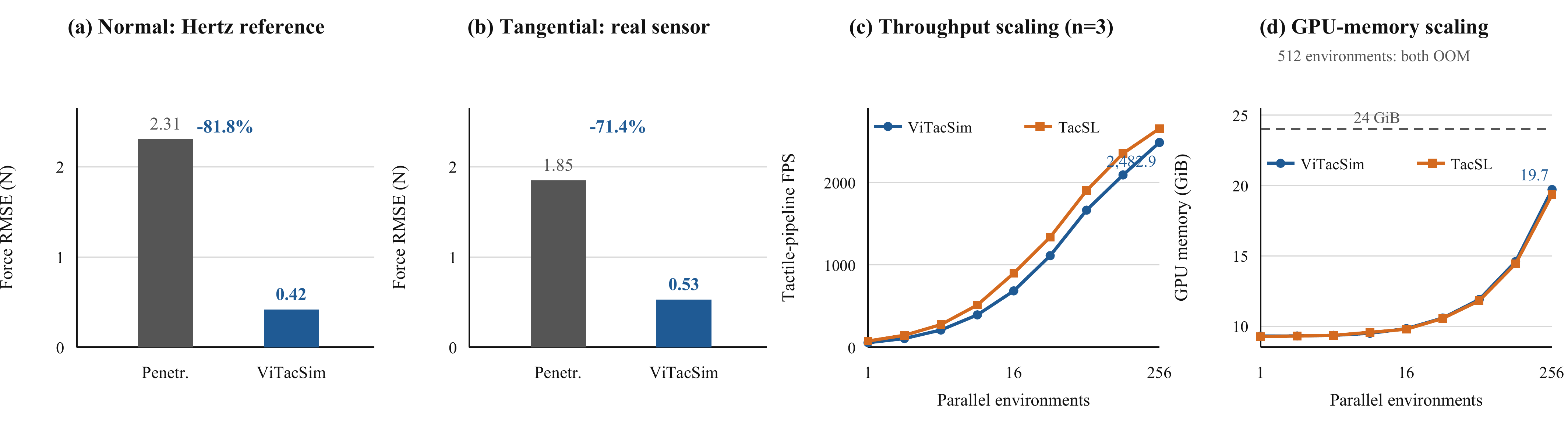}
\caption{ViTacSim validation. Force RMSE against \textbf{(a)} the Hertzian
force--indentation reference and \textbf{(b)} force recorded during controlled
real-sensor sliding; lower is better. \textbf{(c)} Tactile-pipeline
throughput, including dense force-field computation and rendering (mean $\pm$
run-to-run variation, $n=3$). \textbf{(d)} Process GPU memory under the same
scaling protocol; both methods exceed 24~GB at 512 environments.}
\label{fig:sim_validation}
\end{figure*}

\textbf{Physical accuracy.}
Against the penetration-only model, normal RMSE decreases from 2.31 to 0.42~N
against the Hertzian force--indentation reference (81.8\%), and tangential RMSE
decreases from 1.85 to 0.53~N against controlled real-sensor sliding
(71.4\%); see \cref{fig:sim_validation}(a,b).
These tests directly evaluate both reconstructed force components instead of
using policy success as a proxy.

\textbf{Wrench consistency.}
The joint projection in \cref{eq:wrench-projection} constrains the integrated
dense field to the sparse PhysX force and moment while retaining local Coulomb
bounds. This construction-level guarantee complements the two independent
force-reference tests; we report it separately from the empirical RMSE.

\textbf{Systems scaling.}
Maintained-contact scenes run on an Intel Core Ultra~9 285K and RTX~4090D
(24~GB), with identical $20\times25\times3$ force fields and
$320\times240\times3$ tactile images. Tactile-pipeline throughput rises from 56.0 to
2,482.9 FPS (44.3$\times$) from 1 to 256 environments, narrowing the TacSL gap
from 26.7\% to 6.4\% (\cref{fig:sim_validation}(c)). At 256 environments,
ViTacSim uses 20,184~MiB, 376~MiB (1.9\%) above TacSL; both exceed 24~GB at 512
(\cref{fig:sim_validation}(d)).

\subsection{Simulation Performance on ViTacBench}

\begin{table}[t]
\centering
\caption{Simulation summary on representative ViTacBench tasks. Success rate (\%).}
\label{tab:sim_summary}
\resizebox{\columnwidth}{!}{
\small
\setlength{\tabcolsep}{4pt}
\renewcommand{\arraystretch}{1.1}
\begin{tabular}{lcccccc}
\toprule
\textbf{Method}
& \textbf{Peg} & \textbf{Gear} & \textbf{Nut} & \textbf{Grasp} & \textbf{Bottle} & \textbf{Blind} \\ & \textbf{Insert} & \textbf{Mesh} & \textbf{Thread} & & \textbf{Open} & \textbf{Grasp} \\
\midrule
DP & 16.0 & 39.5 & \underline{38.5} & 29.0 & 0.0 & 0.0 \\
ADM\_DP & 20.0 & 44.5 & 34.0 & -- & -- & -- \\
ViTacFormer & -- & -- & -- & 31.0 & 2.0 & 1.5 \\
Ours w/o PT & \underline{21.0} & \textbf{72.0} & {38.0} & \underline{82.5} & \textbf{5.0} & \underline{2.0} \\
\midrule
\textbf{Ours} & \textbf{24.0} & \underline{69.0} & \textbf{44.0}
& \textbf{90.0} & \underline{4.5} & \textbf{3.0} \\
\bottomrule
\end{tabular}}
\end{table}

\Cref{tab:sim_summary} reports six representative ViTacBench results on gripper, dexterous, and bimanual manipulation tasks across all three contact regimes.

Full ViTacPolicy is best on four of the six reported tasks and exceeds every
evaluated external baseline, but the no-pretraining variant remains better on
Gear Mesh and Bottle Open. Moreover, all methods reach only $3\%$ or less on
Blind Grasp, indicating that the Level~3 tasks remain difficult under the
current 100-demonstration budget. We therefore treat the visually degraded
results as evidence of benchmark difficulty rather than a solved regime.

\subsection{Ablations}

Jointly removing SPC and VQ-VAE pretraining lowers Peg, Gear, Nut, and Grasp
from 24/68/44/90\% to 10/27.5/35.5/80\%, supporting their combined role.

\begin{table}[t]
\centering
\caption{Matched tactile-pretraining study under a separate controlled
protocol (success rate in \%).}
\label{tab:alignment_ablation}
\small
\setlength{\tabcolsep}{5pt}
\begin{tabular}{lccc}
\toprule
\textbf{Variant} & \textbf{Gear} & \textbf{Bottle} & \textbf{Peg (L3)} \\
\midrule
Image-only VQ & 13 & 16 & 4 \\
Force-only VQ & 14 & 18 & 5 \\
Ours w/o PT & 17 & 21 & 6 \\
Full shared-codebook & \textbf{18} & \textbf{24} & \textbf{12} \\
\bottomrule
\end{tabular}
\end{table}

The added controls in \cref{tab:alignment_ablation} separate unimodal
pretraining from the shared-codebook configuration. Full has the highest point
estimate on all three tasks, with gains of 4, 6, and 7 points over the stronger
unimodal VQ variant on Gear, Bottle, and L3 Peg. This controlled study uses a
separate evaluation protocol, so its absolute rates are not compared with
\cref{tab:sim_summary}. A separate data-scaling study
raises Full's L3 Peg success from 6\% with 100 demonstrations to 22\% with 500,
showing that this regime is learnable but under-sampled at the default budget.

\subsection{Real-World Experiments}

\textbf{Main results.} We deploy ViTacLab on parallel-gripper,
dexterous-hand, and bimanual platforms for tight-tolerance insertion, grasping,
and coordinated manipulation. The quantitative transfer study in
\cref{tab:sensor_gen} focuses on three gripper tasks whose conditions share the same real-world
training and evaluation protocol; the other deployments remain
qualitative and are not used to support a statistical claim.

\begin{table}[t]
\centering
\caption{Real-world gripper transfer (50 trials per task). Entries are
successes/50 (success rate in \%).}
\label{tab:sensor_gen}
\resizebox{\columnwidth}{!}{
\begin{tabular}{lcccc}
\toprule
\textbf{Variant}
& \textbf{Peg} & \textbf{Nut} & \textbf{Gear} & \textbf{Avg} \\
\midrule
Penetration-only & 10 (20) & 30 (60) & 27 (54) & 44.7 \\
Ours w/o PT & 11 (22) & 33 (66) & 29 (58) & 48.7 \\
Full & \textbf{19 (38)} & \textbf{40 (80)} & \textbf{33 (66)} & \textbf{61.3} \\
\bottomrule
\end{tabular}}
\end{table}

\textbf{Simulation-pretrained tactile transfer.}
To evaluate the transferability of tactile representations learned in ViTacSim,
we pretrain the tactile encoder in simulation and fine-tune it during real-world
policy training. Optical tactile pretraining randomizes lighting and injects
Gaussian marker-displacement noise to reduce sensor-specific appearance bias.
\Cref{tab:sensor_gen} compares the transfer conditions. Full has the highest
point estimate on all three tasks. Its exact 95\% Clopper--Pearson intervals
are [24.7,52.8], [66.3,90.0], and [51.2,78.8]\% for Peg, Nut, and Gear.
Per-task two-sided
Fisher tests against Ours w/o PT give $p=0.126$, $0.176$, and $0.537$. A
task-stratified Cochran--Mantel--Haenszel analysis gives a common odds ratio of
1.82 (95\% CI [1.11,2.98], $p=0.018$), supporting an aggregate benefit without
implying significance on every task. In a separate Peg contribution split,
real-only training obtains 10/50 (20\%), whereas simulation pretraining followed
by real fine-tuning obtains 19/50 (38\%), an 18-point absolute increase.

\section{Conclusion}
\label{sec:conclusion}

We presented \emph{ViTacLab}, a full-stack visuo-tactile framework for
contact-rich robot manipulation. \emph{ViTacSim} anchors tri-modal tactile
generation to PhysX contact-solver outputs, producing
$T_{\text{sparse}}$, $T_{\text{dense}}$, and $T_{\text{img}}$ with corrected
normal responses and decoupled slip--stick tangential reconstruction.
\emph{ViTacBench} provides a hierarchical benchmark across parallel grippers,
dexterous hands, and bimanual systems, with tasks organized from
vision-dominant to tactile-dominant regimes. \emph{ViTacPolicy} aggregates
variable tactile layouts through pose-conditioned Transformer fusion and aligns
$T_{\text{dense}}$ and $T_{\text{img}}$ via shared-codebook VQ-VAE
pretraining. Controlled force tests reduce normal/tangential RMSE by
81.8/71.4\%, the joint projection preserves the sparse solver wrench by
construction, and tactile-pipeline throughput reaches 2,482.9 FPS at 256 environments
with 1.9\% more memory than TacSL. Matched unimodal ablations, five-seed
simulation, and 50-trial real-robot evaluations support the representation and
aggregate transfer gains, while per-task transfer uncertainty remains.

\paragraph{Limitations and future work.}
ViTacLab does not yet fully model deformable objects, adhesion, wear, or
large elastomer deformation, and assumes contacts remain within active tactile
patches. Future work will extend ViTacLab toward richer soft-contact models,
broader real-world tasks, and tactile alignment from unpaired real sensor data.
The Hertzian-reference indentation and real-sensor sliding tests quantify
component-level force error,
but do not yet measure dense spatial-field error, tactile-image fidelity,
numerical wrench residuals across diverse contact distributions, or controlled
parameter sensitivity. Broader physical trials and higher-budget Level~3
studies remain important next steps.

\bibliographystyle{IEEEtran}
\bibliography{main}
\end{document}